%% file: main.tex
\documentclass[10pt,twocolumn,letterpaper]{article}

\usepackage[pagenumbers]{cvpr} 

\input{preamble}
\definecolor{cvprblue}{rgb}{0.21,0.49,0.74}
\usepackage[pagebackref,breaklinks,colorlinks,allcolors=cvprblue]{hyperref}

\usepackage[table]{xcolor}
\definecolor{myviolet}{RGB}{240,226,240}
\definecolor{myviolet1}{RGB}{233,212,233}
\definecolor{myviolet2}{RGB}{248,240,248}

\usepackage{marvosym}
\usepackage{booktabs}
\usepackage{multirow}
\usepackage{graphicx} 
\usepackage{capt-of} 
\usepackage{enumitem}
\usepackage{amsmath}

\title{V-RGBX: Video Editing with Accurate Controls over Intrinsic Properties}

\author{
Ye Fang$^{1,2,*}$, Tong Wu$^{3}$\textsuperscript{\Letter}, Valentin Deschaintre$^{2}$, Duygu Ceylan$^{2}$,
Iliyan Georgiev$^{2}$, \\[0.2em] Chun-Hao Paul Huang$^{2}$, Yiwei Hu$^{2}$,
Xuelin Chen$^{2}$, Tuanfeng Yang Wang$^{2 }$\textsuperscript{\Letter}\\[0.3em]
$^{1}$Fudan University \quad
$^{2}$Adobe Research \quad
$^{3}$Stanford University
}

\begin{document}




\input{sec/0_abstract}

\input{sec/1_intro}
\input{sec/2_relatedwork}

\input{sec/3_method}
\input{sec/4_experiment}
\input{sec/5_discussion}

\input{sec/6_sup}
\clearpage

 {
     \small
     \bibliographystyle{ieeenat_fullname}
     \bibliography{main}
 }



\end{document}

%% file: sec/0_abstract.tex
\twocolumn[{%
\renewcommand\twocolumn[1][]{#1}%
\maketitle
\vspace{-9mm}
\begin{center}
\centering
\includegraphics[width=\textwidth]{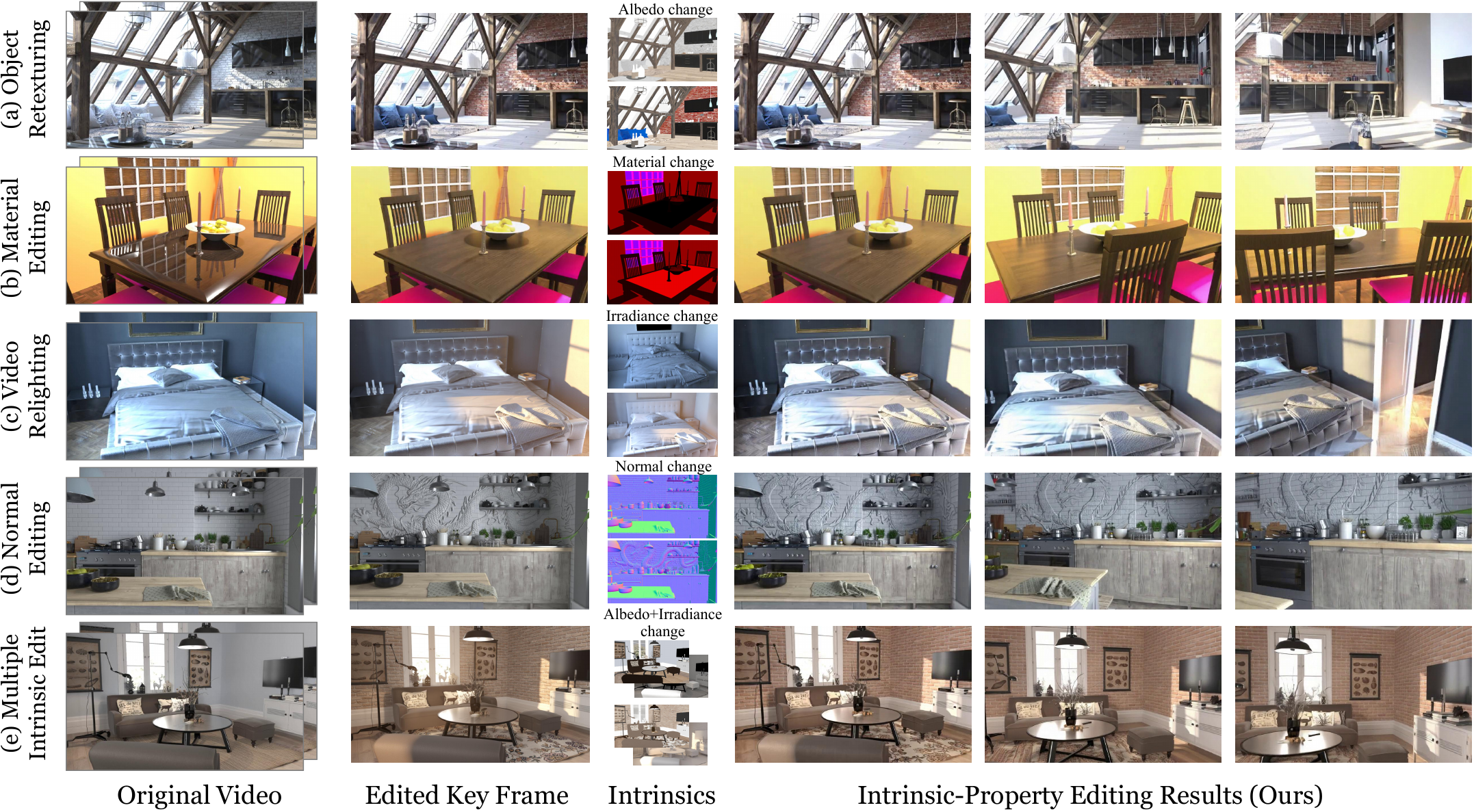}
\label{fig_1:teaser}
\vspace{-4mm}
\captionsetup{hypcap=false} 
\captionof{figure}
{\textbf{Overview.}
Given a source input video and an edited keyframe obtained by manipulating various intrinsic properties, V-RGBX generates an edited video which propagates the edit in an intrinsic aware manner. V-RGBX is an end-to-end framework that understands intrinsic scene properties and uses them for generation to support tasks such as object retexturing, relighting, or material editing, etc.
}
\label{fig1_teaser}
\end{center}
}]

\begingroup
\renewcommand\thefootnote{}
\footnotetext{$^*$This work was partially done while Ye was an intern at Adobe.}

\endgroup

\begin{abstract}
Large-scale video generation models have shown remarkable potential in modeling photorealistic appearance and lighting interactions in real-world scenes.
However, a closed-loop framework that jointly understands intrinsic scene properties (e.g., albedo, normal, material, and irradiance), leverages them for video synthesis, and supports editable intrinsic representations remains unexplored.
We present V-RGBX, the first end-to-end framework for intrinsic-aware video editing. V-RGBX unifies three key capabilities: (1)~video inverse rendering into intrinsic channels, (2)~photorealistic video synthesis from these intrinsic representations, and (3)~keyframe-based video editing conditioned on intrinsic channels.
At the core of V-RGBX is an interleaved conditioning mechanism that enables intuitive, physically grounded video editing through user-selected keyframes, supporting flexible manipulation of any intrinsic modality.
Extensive qualitative and quantitative results show that V-RGBX produces temporally consistent, photorealistic videos while propagating keyframe edits across sequences in a physically plausible manner. We demonstrate its effectiveness in diverse applications, including object appearance editing and scene-level relighting, surpassing the performance of prior methods. Our project website is at: \href{https://aleafy.github.io/vrgbx/}{https://aleafy.github.io/vrgbx}.


\end{abstract}

%% file: sec/1_intro.tex
\section{Introduction}
\label{sec:intro}


Editing captured video sequences using large-scale generative models has witnessed remarkable progress in recent years. With the advent of powerful text-to-video and image-to-video diffusion models, users can now manipulate object appearance, scene layout, and motion dynamics through high-level language or visual instructions \cite{bar2024lumiere, liu2025generative, wang2024motionctrl}. These advances have significantly expanded the scope of creative video synthesis and controllable generation.

However, direct control over intrinsic properties, such as albedo, irradiance, material, and shading, remains largely unexplored. These properties govern the physical realism and consistency of visual appearance, and fine-grained control over them is essential for many downstream applications, including relighting, material editing, and stylized video generation. For instance, one may wish to modify the shadeless texture of an object while preserving its lighting 
or change the environmental lighting without altering surface material. Achieving such effects requires explicit disentanglement of intrinsic factors from RGB observations and the ability to propagate intrinsic edits consistently across time — two capabilities absent in current video generation systems.

Despite the success of video diffusion models in controllable generation, existing approaches lack an explicit intrinsic representation space. Most prior works focus on appearance-level editing (e.g., texture or style transfer) or employ implicit latent conditioning that entangles illumination and material cues. Methods such as GenProp~\cite{liu2025generative}, VACE~\cite{vace}, DaS~\cite{gu2025diffusion}, or AnyV2V~\cite{ku2024anyvv} extend diffusion models with auxiliary modality guidance (e.g., appearance, depth, flow, or semantics). However, these cues are injected directly into the pixel space, without disentanglement in the intrinsic domain. As a result, such approaches often fail to preserve key intrinsic properties consistently across frames after editing.
Moreover, existing solutions typically apply conditioning in a global manner, based on a text prompt or a single reference frame, limiting their flexibility in practical video editing scenarios where multiple, localized edits may occur across different intrinsic modalities and time segments.





To address these limitations, we propose \textbf{V-RGBX}, an intrinsic-space video editing technique that enables keyframe-level control and propagation of intrinsic properties within a generative video model. Our framework begins with an intrinsic decomposition module (video RGB$\rightarrow$X) that extracts physically interpretable modalities from input RGB frames, such as albedo, irradiance, normal, and depth. These modalities constitute a structured intrinsic editing space, where users can selectively modify any modality (e.g., changing albedo color or adjusting illumination) on sparse keyframes. At the core of our system lies a multi-modality conditioning diffusion transformer (DiT) that integrates intrinsic modalities within a unified generation process (video X$\rightarrow$RGB). Rather than conditioning on predefined motion signals (e.g., DaS~\cite{gu2025diffusion}), our model learns temporal dynamics from arbitrary modality inputs in an interleaving manner and propagates sparse keyframe edits throughout the sequence. Untouched modalities are randomly provided and preserved during generation, ensuring temporal and spatial consistency. Such a setup enables reliable multi-touch edits as user can provide multiple touches for different modalities over the sequence.



As shown in \cref{fig1_teaser}, this design enables a broad range of novel video editing capabilities with keyframe level control, including physically grounded relighting, albedo manipulation, and geometry-aware object insertion. Our approach achieves significant improvements in both fidelity and controllability over existing appearance-only baselines and establishes a foundation for physically consistent video generation beyond the RGB domain.
Our main contributions are summarized as follows:

\vspace{3pt}
\begin{itemize}[leftmargin=2em, itemsep=3pt]
    \par\item We introduce V-RGBX, the first end-to-end intrinsic-aware video editing framework, which unifies video inverse rendering into intrinsic channels (video RGB$\rightarrow$X), photorealistic video synthesis from intrinsic representations (video X$\rightarrow$RGB), and keyframe-based video editing conditioned on intrinsic channels.
    \item We propose an interleaving conditioning mechanism for the DiT-based video model, enabling flexible intrinsic conditioning in both video X$\rightarrow$RGB synthesis and keyframe-edit propagation.
    \item We present an intuitive video editing workflow that maintains temporal coherence while supporting keyframe-level, multi-modality control and propagation of intrinsic edits. We demonstrate its effectiveness through intrinsic-driven object appearance editing and scene-level relighting applications.
\end{itemize}

%% file: sec/2_relatedwork.tex
\section{Related work}
\label{sec:RelatedWork}

\subsection{Video Diffusion Models}
Diffusion models have become the leading paradigm for visual synthesis, surpassing GANs in fidelity and diversity by modeling complex data distributions via iterative denoising~\cite{Lee2018StochasticAV,Mathieu2015DeepMV,Tulyakov2017MoCoGANDM,Vondrick2017GeneratingTF,Dhariwal2021DiffusionMB}. 
Building upon their success in image generation, recent works have naturally extended diffusion models to video generation~\cite{ho2022video,he2022latentvideodiffusion,singer2022makeavideo, guo2023animatediff, blattmann2023svd, yang2024cogvideox, wan2025, Sora, Veo}.
Early video diffusion models~\cite{ho2022video, he2022latentvideodiffusion,singer2022makeavideo, blattmann2023align,guo2023animatediff,blattmann2023svd} adopt U-Net~\cite{ronneberger2015unet}-based architectures, incorporating temporal modules to capture frame-to-frame dependencies while maintaining spatial feature processing.
More recent approaches~\cite{yang2024cogvideox,kong2024hunyuanvideo,wan2025} transition to diffusion transformers (DiT), leading to enhanced temporal coherence and overall visual quality.
In this work, we build on the WAN model~\cite{wan2025}, leveraging its strong prior and adapting them for intrinsic properties extraction and video editing.

\subsection{Intrinsic-Aware Diffusion Models}

There has been increasing focus on intrinsic image decomposition, composition, and editing.
Existing methods~\cite{zeng2024rgb, dirik2025prism, Sartor:2025:TCD} enables simple edits by modifying intrinsic components and recomposing them, while IntrinsicEdit~\cite{lyu2025intrinsic} introduces a short residual optimization step that allows precise and controllable intrinsic attribute editing.
However, these methods operate purely at the image level. When extended to videos, the introduction of the temporal dimension raises new challenges: edits applied to a single frame must be accurately propagated to subsequent frames while maintaining temporal consistency. DiffusionRenderer~\cite{DiffusionRenderer} enables video decomposition and re-composition but doesn't enable propagation of pixelwise editing.
X2Video~\cite{huang2025x2video} is a video-level X→RGB method that mainly extends image-level models to the temporal domain, while it cannot directly propagate or edit intrinsic attributes frame by frame without a complete sequence of edited intrinsic properties.

%
To address this, our method builds upon a pretrained video diffusion framework~\cite{wan2025} and proposes a single-frame interleaved conditioning strategy to fuse and propagate intrinsic controls across time.
Moreover, a type-aware embedding is introduced to explicitly differentiate between various intrinsic types at each timestep, ensuring precise edits while achieving cross-frame consistency and stability in intrinsic-aware video synthesis.

\subsection{Controlled Video Generation and Editing}
Controlled video generation conditioned on various signals has recently attracted growing attention due to its wide range of applications.
For example, a series of works have explored camera control~\cite{wang2024motionctrl,he2025cameractrl,bahmani2025vd3d,bahmani2025ac3d,bai2025syncammaster,bai2025recammaster}, while other methods incorporate structural conditioning via point clouds, object tracking, or 3D-aware priors, which have proven effective in improving spatial consistency and trajectory alignment~\cite{yu2024viewcrafter,gu2025das,yu2025trajectorycrafter}.
Beyond spatial control, several models have begun supporting action-based or scene-level conditioning~\cite{parkerholder2024genie2,oasis2024,wu2025spmem}, collectively advancing video generation toward a world simulator paradigm. 
Recent advances have also enabled general video editing~\cite{liu2025generative,vace}, supporting multiple conditioning modalities with different editing tasks in a unified framework. 
However, achieving more physically realistic rendering and editable generation remains challenging, as intrinsic-based tasks are still largely unexplored. 
In contrast, our framework can recover multiple intrinsic properties, including illumination, from video inputs and re-compose them through intrinsic-conditioned video generation. It thus provides a complete editing pipeline of RGB→X and X→RGB, allowing precise propagation and control of per-frame intrinsic edits across time.


%% file: sec/3_method.tex
\section{Method}
\label{sec:Method}

\begin{figure*}[t]
  \centering
  \vspace{-6pt}
  \includegraphics[width=\linewidth]{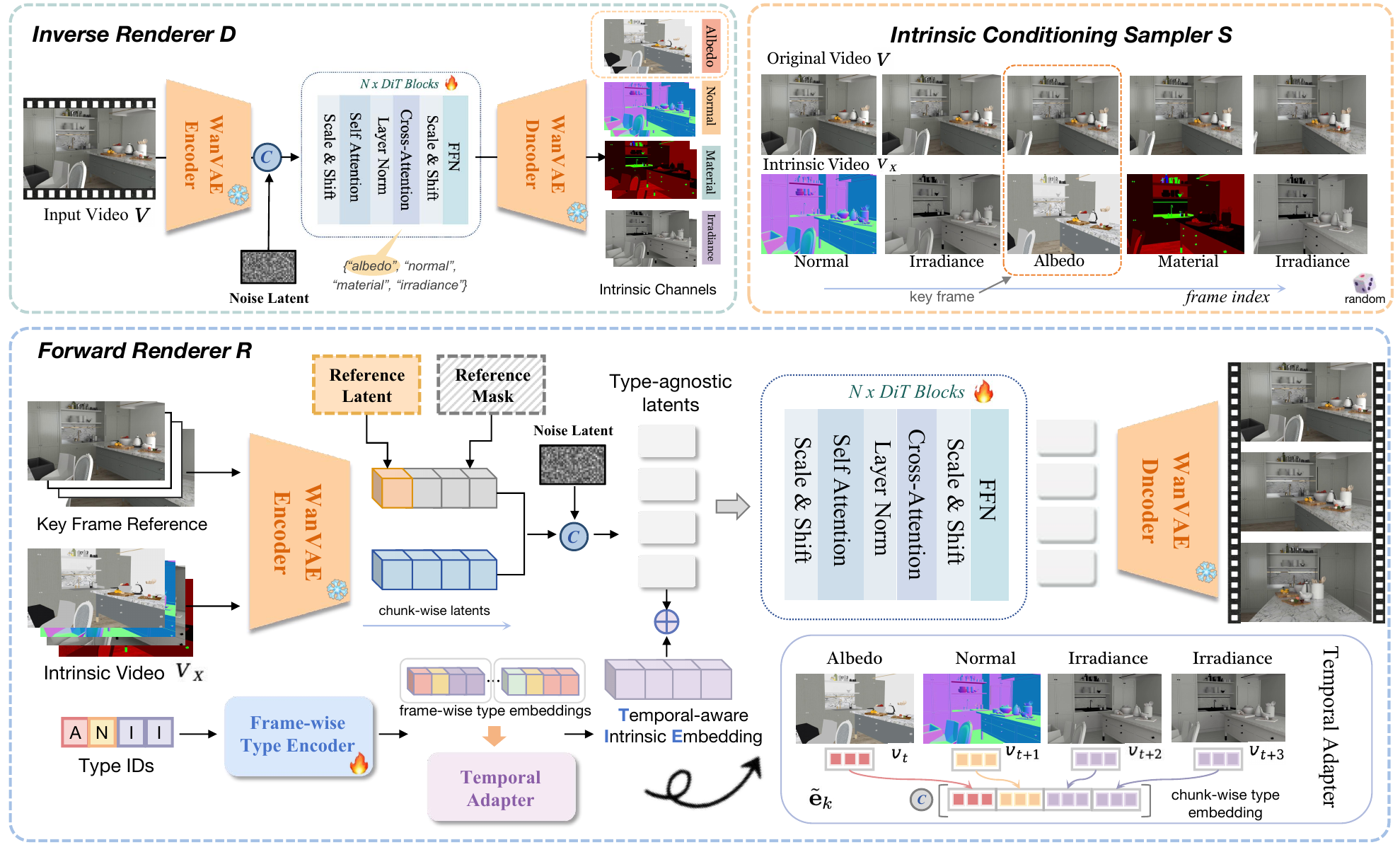}
  \caption{\textbf{The overall architecture of V-RGBX.} Our framework consists of three parts: (1) Inverse Renderer $D$, which decomposes the input video into albedo, normal, material, and irradiance channels; (2) Intrinsic Conditioning Sampler $S$, which interleaves edited keyframe intrinsics with non-conflicted random intrinsic frames to form a unified intrinsic conditioning video; and (3) Forward Renderer $R$, which integrates the intrinsic video, keyframe reference, and temporal-aware intrinsic embeddings to synthesize the output RGB video and consistently propagate intrinsic properties across time.}
  \label{fig:method}
  \vspace{-5mm}
\end{figure*}


\subsection{Overview}

We propose \textbf{V-RGBX}, an intrinsic-aware video editing framework that enables precise and temporally consistent propagation of intrinsic edits. 
Unlike previous video diffusion models trained purely in the RGB domain, which often lead to entangled control over lighting, texture, and geometry, \textbf{V-RGBX} introduces an explicit \textit{intrinsic-conditioned representation space}, ensuring that edits to one property do not unintentionally affect others.

Specifically, starting with an input video $V=\{v_1, v_2, \ldots, v_T\}$, the user selects a set of keyframes $\{v'_{i_1}, \ldots, v'_{i_k}\}$ and performs 
edits using Photoshop~\cite{adobe2025photoshop} or a text-to-image generator~\cite{nanabanana2025,rombach2022high}. 
Image-space inverse rendering techniques~\cite{zeng2024rgb, Luo2024IntrinsicDiffusion} are then applied to identify which intrinsic modalities have been modified. 
Our goal is to propagate these edits across time, generating an edited video sequence 
$V'=\{v'_1, v'_2, \ldots, v'_T\}$, while preserving the untouched intrinsic properties and maintaining temporal coherence.

As illustrated in Fig.~\ref{fig:method}, \textbf{V-RGBX} consists of three main components: 
(1)~(RGB$\rightarrow$X) Starting from sequential RGB frames, an \textbf{inverse rendering model} $D(\cdot)$ estimates the corresponding intrinsic channels, such as \textbf{A}lbedo, \textbf{N}ormal, \textbf{M}aterial, and \textbf{I}rradiance, $D(V) = \{V_A, V_N, V_M, V_I\} \in \mathbb{R}^{T \times 3 \times H \times W}$, where $T$, $H$, and $W$ denote the number of frames, height, and width of the video, respectively. The material channel consists of surface attributes such as roughness, metallic, and ambient occlusion;
(2)~After keyframe edits are applied, the modified intrinsic modalities of neighboring frames in $D(V)$ can no longer be directly used for conditioning during video generation. To address this, we combine the edited intrinsic channels at the keyframes with randomly interleaved, untouched intrinsic channels from other frames to construct a streamed \textit{intrinsic conditioning} sequence $V_X' = \text{Sample}(D(V))$;
(3)~(X$\rightarrow$RGB) At the core of our framework, a \textbf{forward rendering network} $R(\cdot)$ synthesizes the output video conditioned on both the streamed intrinsic conditioning and the edited keyframes $V' = R(\{v'_{i_1}, \ldots, v'_{i_k}\}, V_X')$.

\subsection{
\texorpdfstring{Inverse Rendering (RGB$\rightarrow$X)}%
{Inverse Rendering (RGB to X)}
}


The inverse rendering model $D(\cdot)$ predicts intrinsic channels from the input video $V=\{v_1,\ldots,v_T\}$, including $\mathsf{albedo}$, $\mathsf{normal}$, $\mathsf{material}$, and $\mathsf{irradiance}$. 
We adopt a Diffusion Transformer (DiT) backbone~\cite{wan2025} and condition the denoising process with 
$h_t = [x_t^z \Vert \mathcal{E}(V)]$, 
where $x_t^z$ denotes the initial noisy latent, $\mathcal{E}(\cdot)$ is the frozen Wan-VAE encoder, and $\Vert$ represents channel-wise concatenation. 
The target modality name is encoded as a text prompt using CLIP embeddings~\cite{radford2021learning}. 
We fine-tune the backbone with the velocity-prediction objective~\cite{peebles2023scalable} for improved training stability. 
The denoised latent is finally decoded by the frozen Wan-VAE decoder into a three-channel intrinsic image corresponding to the target modality.



\subsection{Intrinsic-aware Conditioning}

After keyframe edits are applied, the unedited intrinsic modalities of neighboring frames in $D(V)$ can no longer be directly used for conditioning during video generation, as they may introduce conflicts with the edited keyframes. 
Previous approaches (e.g., GenProp~\cite{liu2025generative} and VACE~\cite{vace}) handle such cases by inserting empty tokens to compensate for missing frames in the conditioning sequence. 
However, this strategy leads to substantial memory overhead in our setting, where multiple intrinsic channels are jointly modeled, and limits scalability to additional intrinsic modalities. We propose to ensure temporal coherence and cross-modality consistency by interleaving the decomposed intrinsic channels into a single conditioning sequence:
\[
V'_X = \text{Sample}\!\left(\{V_A, V_N, V_M, V_I\}\right) = \{v_1^x, v_2^x, \ldots, v_T^x\},
\]
where $\text{Sample}(\cdot)$ denotes a temporal multiplexing operation that alternates intrinsic modalities over time. 

In the intrinsic-driven editing setting, we assume that one or more intrinsic modalities, denoted as $\mathcal{M}_t$, are modified at keyframe $t$. 
When sampling the conditioning signals for edited keyframes, we randomly draw from the edited modalities $\mathcal{M}_t$, while for the remaining frames, we sample from the \textit{non-conflicted} intrinsic modalities of that frame. 
A modality is considered \textit{conflicted} if it is affected by user edits in any keyframe, as its altered content may introduce inconsistencies when used as conditioning input. 
Formally,
\[
v_t^x = \!
\begin{cases}
\small\text{RandomSample}\!\left(\mathcal{M}_t\right), & \!\!\! t \in \{v'_{i_1},..., v'_{i_k}\}, \\[4pt]
\small\text{RandomSample}\!\left(\{\!A,\!N,\!M,\! I\} \! \setminus \! \mathcal{K}_t\right), & \!\!\! \small\text{otherwise.}
\end{cases}
\]

This design encourages cross-modality propagation and enhances temporal stability. 
It naturally adapts to diverse attribute combinations and incomplete inputs. This provides a lightweight and extensible framework for intrinsic-aware video generation and editing applications.

\subsection{Forward Rendering with Intrinsic-aware Conditioning}

Given the edited RGB keyframes and the composed intrinsic conditioning sequence, our forward rendering model synthesizes a temporally coherent video that faithfully reflects the keyframe edits while preserving the intrinsic properties encoded in the conditioning inputs. 
We build upon WAN~\cite{wan2025}, a large-scale video generation backbone. Our model integrates the intrinsic conditioning sequence in a unified and interleaved sequence modulated with a modality-aware embedding and uses an edited keyframe for further conditioning.

\paragraph{Keyframe referencing.} 
When keyframes are edited in the RGB domain, the edited keyframes can serve as visual guidance alongside the intrinsic video conditioning. 
To enable this, we align their temporal length with the original video sequence by extending the keyframes with empty tokens.


The extended sequence, $\Sigma$, is then processed by the Wan-VAE encoder to obtain a reference latent that aligns in shape with the backbone latent space.
We concatenate the noisy latent $x_t^z$, the embedding of the conditioning interleaved intrinsic sequence, $\mathcal{E}_{\text{VAE}}(V'_X)$, and the embedding of reference sequence $\mathcal{E}_{\text{VAE}}(\Sigma)$ along the channel dimension to compose the input to the diffusion model. Specifically,
\begin{equation}
    \mathbf{z}_t = [x_t^z \Vert \mathcal{E}_{\text{VAE}}(V'_X) \Vert \mathcal{E}_{\text{VAE}}(\Sigma)].
\end{equation}
By encoding the keyframes as reference signals and injecting them jointly with the intrinsic conditioning, the model learns to capture the overall visual content of the scene and compensates for information not explicitly represented in the intrinsic channels.

\paragraph{Temporal-aware intrinsic embedding.}
Following our backbone video generator, the encoder compresses every four consecutive frames into one latent chunk. 
However, these four conditioning frames may correspond to different intrinsic modalities. 
To embed an intrinsic conditioning sequence 
into the temporally compressed latent space of the DiT backbone, we propose a \textbf{T}emporal-aware \textbf{I}ntrinsic \textbf{E}mbedding (TIE) that \textit{packs} per-frame modality embeddings within the chunk dimension, preserving both temporal order and modality identity.


Let each frame $i$ be assigned a modality index $m_i \in \{\mathsf{albedo}, \mathsf{normal}, \mathsf{irradiance}, \mathsf{material}, \ldots\}$, we compute its embedding as
\begin{align}
\mathbf{e}_i = \mathbf{W}\,\phi(m_i), \quad \mathbf{e}_i \in \mathbb{R}^{d},
\end{align}
where $\phi(\cdot)$ is a one-hot modality indicator and the type encoder $\mathbf{W}$ is a learnable embedding matrix.
Similar to~\cite{wan2025}, we then construct a packed embedding for each latent chunk $k$ via a temporal adapter,
\begin{align}
\tilde{\mathbf{e}}_k =
\begin{cases}
[\mathbf{e}_1 \Vert \mathbf{e}_1 \Vert \mathbf{e}_1 \Vert \mathbf{e}_1], & k = 1, \\[3pt]
[\mathbf{e}_{4k-3} \Vert \mathbf{e}_{4k-2} \Vert \mathbf{e}_{4k-1} \Vert \mathbf{e}_{4k}], & k > 1,
\end{cases}
\quad
\tilde{\mathbf{e}}_k \in \mathbb{R}^{4d},
\end{align}
where $\Vert$ denotes concatenation along the feature dimension.

After patchification, each latent chunk $\mathbf{z}_t^k \in \mathbb{R}^{H' \times W' \times 4d}$ is modulated by its corresponding packed embedding:
\begin{align}
\tilde{\mathbf{z}}_t^k = \mathbf{z}_t^k + \gamma\,\tilde{\mathbf{e}}_k^{*},
\end{align}
where $\tilde{\mathbf{e}}_k^{*}$ represents the spatial broadcasting of the modality embedding, and $\gamma$ is a constant scaling factor empirically set to 1. 
This formulation enables each latent chunk to carry explicit modality information while maintaining temporal order within the chunk.

\paragraph{Training and Inference.}
Following the inverse rendering stage, we adopt the \textit{v}-prediction objective~\cite{peebles2023scalable} for training. 
For simplicity, text conditioning is omitted in this setting. 
During training, the keyframe reference is randomly dropped with a probability of $p_\mathrm{drop}=0.3$. 
At inference time, classifier-free guidance~\cite{ho2022classifier} is applied to the reference conditioning to balance fidelity and edit consistency.



%% file: sec/4_experiment.tex
\section{Experiments}
\label{sec:Experiments}

\begin{figure*}[t]
  \centering
  \includegraphics[width=\linewidth]{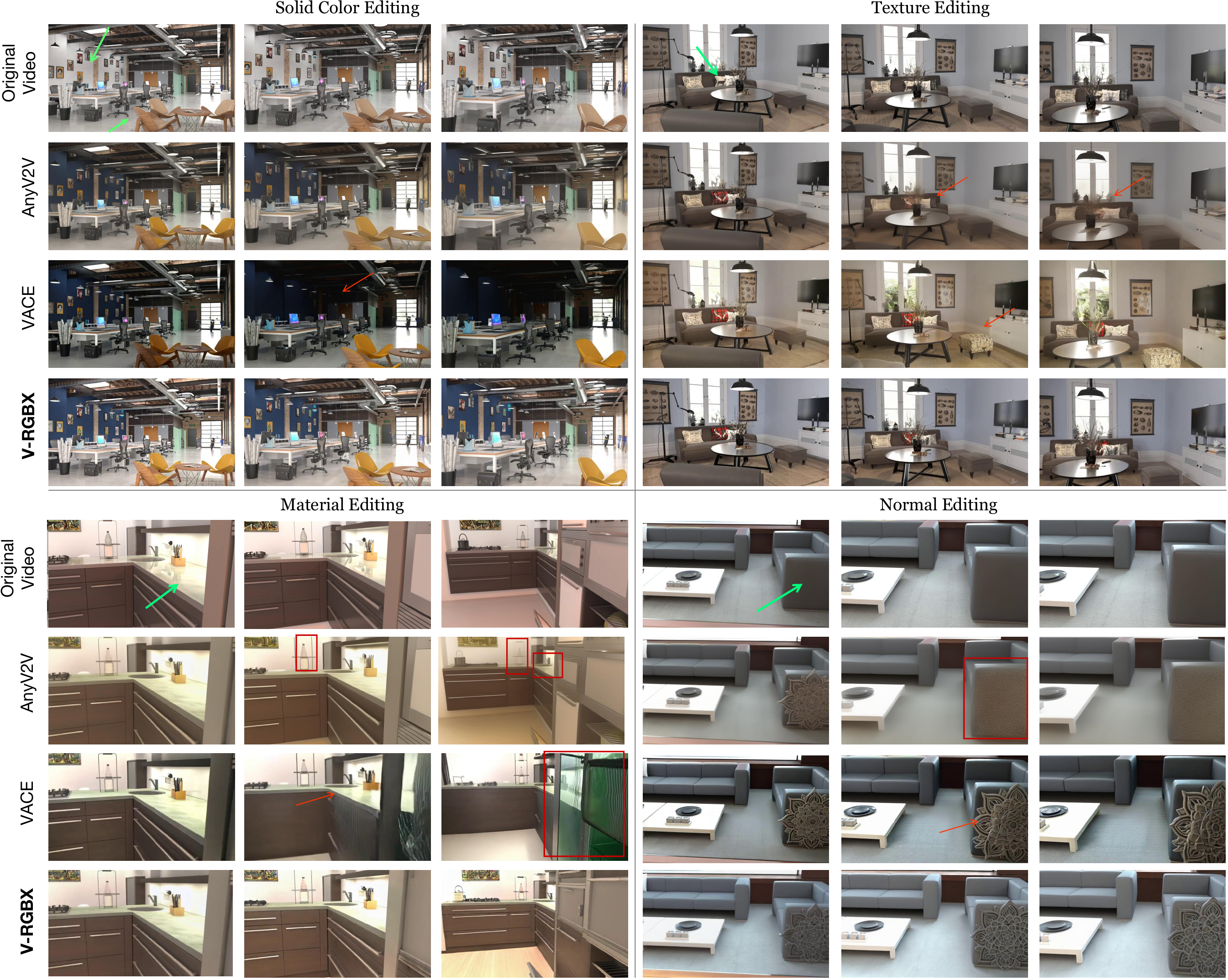}
  \caption{
  \textbf{Qualitative comparison on intrinsic-aware editing.} 
    We show the results of keyframe-edit video propagation based on a single intrinsic channel. AnyV2V\cite{ku2024anyv2v} and VACE\cite{vace} suffer from property drifting, exhibiting unexpected evolution over time, and fail to achieve accurate disentanglement among different intrinsic channels. In contrast, our method consistently propagates the intrinsic-aware edits throughout the sequence. (Green arrows mark the edited regions in the original video’s keyframe, whereas red arrows and boxes denote artifacts generated by the baseline methods.)
  }
  \label{fig:material editing}
  \vspace{-5mm}
\end{figure*}

\begin{figure*}[t]
  \centering
  \includegraphics[width=\linewidth]{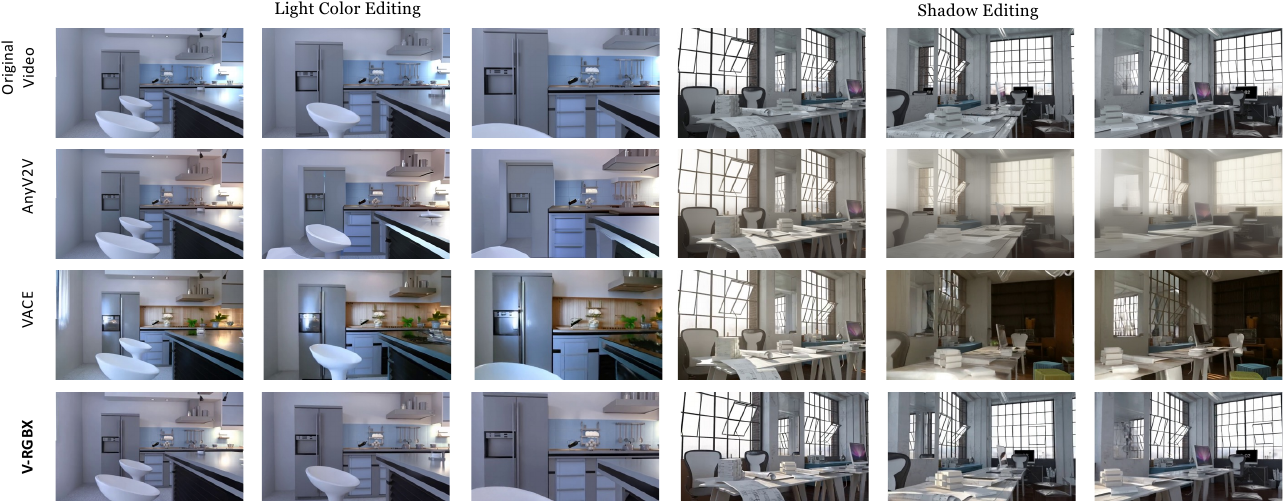}
  \caption{
  \textbf{Qualitative comparison on relighting.} 
  We evaluate our method on light color and shadow editing tasks. AnyV2V exhibits geometry and appearance drifting as generation progresses, while VACE struggles to disentangle lighting effects, leading to unintended changes in other channels and even introducing new assets. Our approach naturally performs relighting in both tasks and produces results that closely match the ground truth.
  }
  \label{fig:relighting}
\end{figure*}

We extensively evaluate \textbf{V-RGBX} on a diverse set of synthetic and real-world datasets. 
Details of our experimental setup are provided in Sec.~\ref{exp_details}. 
We conduct comprehensive comparisons and ablation studies across three core tasks: 
(\textbf{RGB}$\rightarrow$\textbf{X}) \textit{inverse rendering} (Sec.~\ref{exp_rgb2x}), 
(\textbf{X}$\rightarrow$\textbf{RGB}) \textit{intrinsic-aware video synthesis} (Sec.~\ref{exp_x2rgb}), 
and \textit{keyframe-based video editing} (Sec.~\ref{exp_conditioning}). 
Finally, we demonstrate the versatility of our framework through a range of \textit{intrinsic-driven video editing applications} in Sec.~\ref{exp_application}.

\subsection{Experiment settings}
\label{exp_details}




We train \textbf{V-RGBX} on an internal synthetic dataset rendered from 127 Evermotion~\cite{zeng2024rgb} interior scenes, producing 171K frames with paired supervision across RGB, albedo, normal, material, and irradiance channels. 
All models are trained at a resolution of $832\times480$ using 32 NVIDIA A100 (80GB) GPUs. 
Both the Inverse Renderer $D$ and Forward Renderer $R$ are initialized from the pretrained Wan~2.1~\cite{wan2025} T2V-1.3B DiT backbone and trained for $27$K and $12$K iterations, respectively, with a learning rate of $2\times10^{-4}$. 
The type encoder $\mathbf{W}$ is trained from scratch.

For evaluation, we select 85 videos from unseen Evermotion scenes and 85 videos from RealEstate10K~\cite{zhou2018stereo} to assess performance on both synthetic and real-world data. 
To maintain consistency with baseline methods, only the first frame of each video is used as the keyframe input by default.





\paragraph{Metrics.}
We evaluate both forward and inverse rendering performance using PSNR, SSIM, and LPIPS~\cite{zhang2018unreasonable}. 
For albedo evaluation, a three-channel scaling factor is applied via least-squares optimization before computing the metrics to account for global scaling ambiguity. 
To assess video generation quality, we use FVD~\cite{unterthiner2019fvd}, and for temporal coherence, we adopt the smoothness score from VBench~\cite{huang2024vbench}.

\subsection{Inverse rendering (RGB$\rightarrow$X)}
\label{exp_rgb2x}
We compare our method against RGB$\leftrightarrow$X~\cite{zeng2024rgb} and DiffusionRenderer~\cite{DiffusionRenderer} for intrinsic channel extraction on our synthetic dataset. 
Since RGB$\leftrightarrow$X is an image-based approach, we perform inverse rendering on each frame independently. 
Consequently, its temporal consistency is not directly comparable to other video-based baselines. DiffusionRenderer does not estimate lighting, so we report results only on the remaining intrinsic channels. 
As shown in Table~\ref{tab:rgb2x}, \textbf{V-RGBX} outperforms other baselines in all intrinsic modalities for frame-wise pixel-aligned accuracy. 
We show qualitative comparisons of the predicted intrinsic channels in supplemental material, showing that our method maintains better temporal coherence than other baselines.

\begin{figure*}[t]
  \centering
  \includegraphics[width=\linewidth]{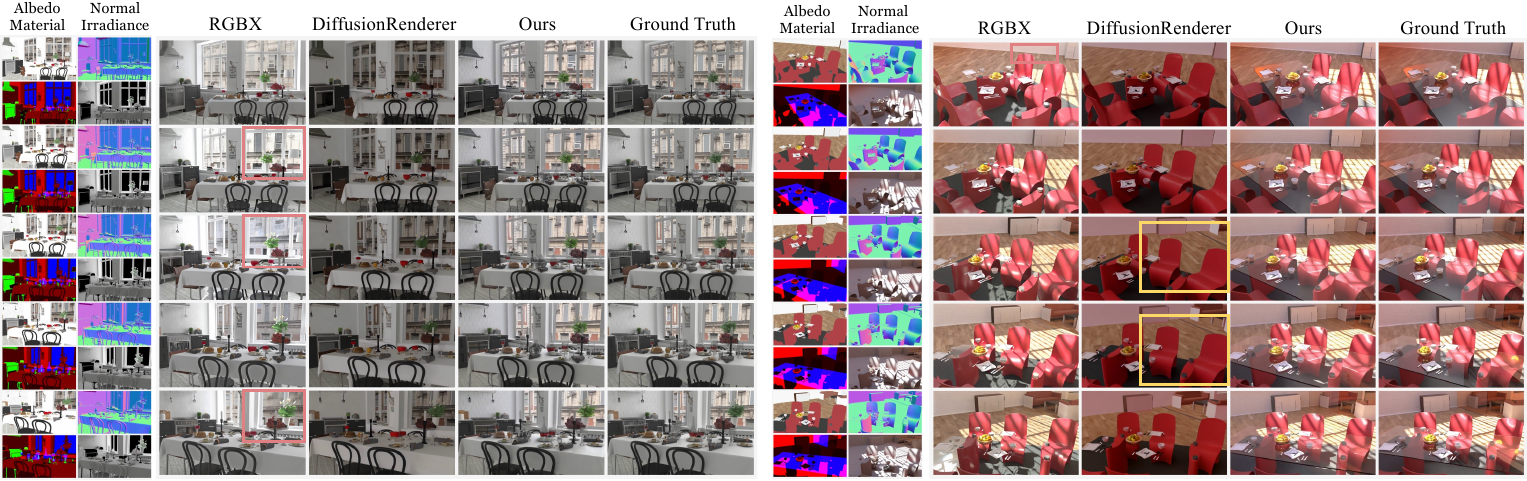}
  \caption{
  \textbf{Qualitative comparisons for the X$\rightarrow$RGB task.} 
  Pink frames highlight temporal or spatial inconsistencies observed in RGBX results, while yellow frames indicate inaccurate or missing shadow modeling produced by DiffusionRenderer. 
  Our method achieves temporal coherence, reliable shadow and light modeling, and overall more faithful scene generation from X to video.
}
  \label{fig:x2rgb}
\end{figure*}

\subsection{Intrinsic-aware video generation (X$\rightarrow$RGB)}
\label{exp_x2rgb}
We first evaluate \textbf{V-RGBX} on the synthetic dataset, where ground-truth intrinsic channels are available. 
Under our interleaved conditioning setting, one intrinsic modality is randomly sampled for each frame, along with a randomly selected RGB frame as the reference image. 
Since our synthetic dataset only provides rendered irradiance maps, we follow~\cite{phongthawee2024diffusionlight} to estimate an environment map from the first frame of each video sequence for fair comparison with DiffusionRenderer~\cite{DiffusionRenderer}. 
As shown in Fig.~\ref{fig:x2rgb}, although the estimated environment maps are reasonable, DiffusionRenderer fails to reproduce realistic reflections in the test scenes, resulting in misleading high smoothness scores (Table~\ref{tab:x2rgb}). 

We further conduct an ablation study to assess the impact of keyframe reference images by comparing V-RGBX variants with and without reference conditioning. 
As evidenced by both quantitative metrics and visual comparisons, incorporating a reference image helps the model better capture the visual style of the scene and compensates for information not explicitly represented in the intrinsic channels. 

As a unified framework, \textbf{V-RGBX} can also be evaluated in an end-to-end manner via cycle consistency. 
On both synthetic and real datasets, we perform inverse rendering followed by forward rendering using our approach and the baseline methods, then compare the generated video sequences with the original inputs. 
Consistent with previous settings, RGB$\leftrightarrow$X~\cite{zeng2024rgb} operates in a frame-wise manner, while DiffusionRenderer uses environment maps estimated by~\cite{phongthawee2024diffusionlight}. 
As shown in Table~\ref{tab:cycle}, \textbf{V-RGBX} achieves the best overall performance among all baselines, demonstrating superior temporal coherence and frame-wise pixel-aligned accuracy on both real and synthetic datasets. 
DiffusionRenderer again reports spuriously high smoothness scores, as its forward-rendered outputs exhibit faded in qualitative evaluations.
Additional qualitative comparisons are provided in the supplementary materials.

\subsection{Keyframe based video conditioning}
\label{exp_conditioning}
Our method supports flexible conditioning through intrinsic channels during inference. 
In this experiment, we evaluate \textbf{V-RGBX} under different conditioning schemes: 
(1) excluding a specific modality $\chi$ from the conditioning sequence, and 
(2) using $\chi$ only in the first frame,
where in both settings, $\chi$ can be either albedo or irradiance. These settings simulate real editing scenarios in which certain modalities cannot be used due to conflicts with the edited keyframes. 
As shown in Table~\ref{tab:x_guided_generation_metrics}, when an RGB reference frame is provided, our trained model achieves quantitatively comparable performance as in Table~\ref{tab:x2rgb} under different conditioning schemes, demonstrating the robustness and reliability of \textbf{V-RGBX} in real intrinsic-guided video editing scenarios.

\subsection{Applications}
\label{exp_application}

\textbf{V-RGBX} enables keyframe-based, intrinsic-aware editing of input videos. 
In this evaluation, we use a text-driven image editing tool, NanoBanana~\cite{nanabanana2025}, to modify selected keyframes. 
These edited keyframes are encoded as references and jointly condition the video generation process together with the intrinsic conditioning sampled from the video's intrinsic channels and keyframe intrinsics. 
In addition to Fig.~\ref{fig1_teaser}, Fig.~\ref{fig:material editing} and~\ref{fig:relighting} present examples of appearance editing and scene relighting tasks, respectively. 
As shown, \textbf{V-RGBX} faithfully incorporates the keyframe edits while preserving the untouched intrinsic content from the original video frames, significantly outperforms the existing baselines. 
We include the input and output videos, edited keyframes, extracted intrinsic channels, and additional intermediate results in the supplementary material.






%


\begin{table}[t]
\caption{
\textbf{Quantitative evaluation on the RGB$\rightarrow$X task.}
We show that our method significantly outperforms previous approaches in terms of albedo, normal, and irradiance estimation.
}
\centering
\small
\setlength{\tabcolsep}{4pt}
\resizebox{\columnwidth}{!}{%
\begin{tabular}{l|cc|cc|cc}
\toprule
Method &
\multicolumn{2}{c|}{\textbf{Albedo}} &
\multicolumn{2}{c|}{\textbf{Normal}} &
\multicolumn{2}{c}{\textbf{Irradiance}} \\
& PSNR$\uparrow$ & LPIPS$\downarrow$
& PSNR$\uparrow$ & LPIPS$\downarrow$
& PSNR$\uparrow$ & LPIPS$\downarrow$ \\
\midrule
RGBX              & 14.04 & 0.2872 & 19.44 & 0.1800 & 11.92 & 0.2994 \\
DiffusionRenderer & 17.40 & 0.3002 & 21.04 & 0.1817 & - & - \\
\rowcolor{myviolet}
V\mbox{-}RGBX (ours) & \textbf{17.73} & \textbf{0.2406} & \textbf{21.59} & \textbf{0.1407}  & \textbf{19.94} & \textbf{0.2187} \\
\bottomrule
\end{tabular}%
}

\label{tab:rgb2x}
\end{table}

\begin{table}[t]

\caption{\textbf{Quantitative evaluation on the X$\rightarrow$RGB task.}
We show that our method substantially surpasses previous approaches by achieving higher forward rendering accuracy, improved video quality, and smoother temporal consistency. Our method is further enhanced when the reference frame is provided.
}
\centering

\resizebox{0.48\textwidth}{!}{
\begin{tabular}{lccccc}
\toprule
\textbf{Method} & \textbf{PSNR}$\uparrow$ & \textbf{SSIM}$\uparrow$ &
\textbf{LPIPS}$\downarrow$ & \textbf{FVD}$\downarrow$ & \textbf{Smoothness}$\uparrow$ \\
\midrule
RGBX       & 16.53 & 0.7154 & 0.2417 & 1037.15 & 0.9469 \\
DiffusionRenderer$^*$ & 12.66 & 0.6475 & 0.3376 & 1015.09 & \textbf{0.9883} \\
\rowcolor{myviolet2}
V-RGBX (w/o ref.) & \underline{21.48} & \underline{0.7908} &
\underline{0.2064} & \underline{401.62} & \underline{0.9814} \\
\rowcolor{myviolet}
V-RGBX (ours) & \textbf{22.42} & \textbf{0.7952} &
\textbf{0.1930} & \textbf{367.89} & 0.9805 \\
\bottomrule
\end{tabular}
}

\label{tab:x2rgb}
\end{table}

\begin{table}[t]
\caption{
\textbf{Quantitative evaluation on RGB$\rightarrow$X$\rightarrow$RGB cycle consistency.}
We conduct comparisons on both synthetic and real-world data, where our approach achieves the best cycle consistency in inverse and forward rendering, demonstrating its robustness and high fidelity.
}
\centering
\small
\resizebox{\columnwidth}{!}{%
\begin{tabular}{l|lcccc}
\toprule
\textbf{Dataset} & \textbf{Method} & \textbf{PSNR}$\uparrow$ & \textbf{SSIM}$\uparrow$ & \textbf{FVD}$\downarrow$ & \textbf{Smoothness}$\uparrow$ \\
\midrule
\multirow{3}{*}{Evermotion} 
 & RGBX              &   15.29   &   0.7539   &   1099.04   &   0.9485   \\
 & DiffusionRenderer$^*$ &   12.42   &   0.6311   &   1073.98   &   0.9803   \\
& \cellcolor{myviolet}V-RGBX (ours)     &   \cellcolor{myviolet}\textbf{22.57}   &   \cellcolor{myviolet}\textbf{0.7985}   &   \cellcolor{myviolet}\textbf{367.61}   &   \cellcolor{myviolet}\textbf{0.9808}   \\
\midrule
\multirow{3}{*}{RealEstate10K} 
 & RGBX           &   14.40     &    0.6411  &    2082.81  &   0.9307   \\
 & DiffusionRenderer$^*$ &   12.53  &   0.6272   &   1643.05   &   0.9839   \\
& \cellcolor{myviolet}V-RGBX (ours)     &    \cellcolor{myviolet}\textbf{17.88}  &   \cellcolor{myviolet}\textbf{0.7533}   &   \cellcolor{myviolet}\textbf{633.76}   &   \cellcolor{myviolet}\textbf{0.9870}   \\
\bottomrule
\end{tabular}}
\label{tab:cycle}
\end{table}


\begin{table}[t]{

\caption{\textbf{Quantitative comparison on different control strategies in the X$\rightarrow$RGB task.}
We drop one intrinsic channel (albedo or irradiance) during the X$\rightarrow$RGB generation and find that our method still demonstrates strong robustness in rendering. When the missing channel is provided only for the first frame as reference, the results show a corresponding improvement, indicating that the model effectively propagates the guidance across the entire sequence.\label{tab:x_guided_generation_metrics}
}
\centering
\scriptsize
\setlength{\tabcolsep}{4pt}
\renewcommand{\arraystretch}{1.2}
\resizebox{0.48\textwidth}{!}{%
\begin{tabular}{ll|ccc|ccc}
\toprule
\textbf{Control Type X} & \textbf{Method}
& \multicolumn{3}{c|}{\textbf{Albedo}} 
& \multicolumn{3}{c}{\textbf{Irradiance}} \\
\cmidrule(lr){3-5} \cmidrule(lr){6-8}
& 
 &PSNR$\uparrow$ &  SSIM$\uparrow$ & FVD$\downarrow$
& PSNR$\uparrow$ & SSIM$\uparrow$ & FVD$\downarrow$ \\
\midrule
\multirow{2}{*}{Drop X channel} 
& Ours (w/o ref) & 17.18 & 0.7236 & 907.63 & 17.43 & 0.7350 & 702.16 \\
& Ours & 20.83 & 0.7623 & 549.71 & 21.70 & 0.7807 &  441.05 \\
\midrule
\multirow{2}{*}{1st-Frame X-Guided} 
& Ours (w/o ref)  & 20.17 & 0.7652 & 496.09 & 20.67 & 0.7775 & 461.53 \\
& Ours & 21.65 & 0.7738 & 427.56 & 21.82 & 0.7844 & 396.40 \\
\bottomrule
\end{tabular}%
}
}

\vspace{-5mm}
\end{table}




%% file: sec/5_discussion.tex
\section{Discussion}

In this paper, we present \textbf{V-RGBX}, an end-to-end framework for video editing with intrinsic-level control. 
While \textbf{V-RGBX} demonstrates strong performance in intrinsic-aware video editing, several limitations remain. 
First, the model is trained only on indoor synthetic datasets and may therefore struggle to generalize to out-of-distribution scenarios, such as outdoor scenes. 
Second, the current intrinsic conditioning samples exactly one modality per frame, which limits its ability to capture complex, multi-attribute edits on keyframes. 
Third, the framework relies on a pretrained video backbone~\cite{wan2025}, constraining scalability in both video length and real-time performance. 
Looking forward, the framework could be extended with long-range generation capabilities~\cite{li2025arlon}, enabling more flexible and persistent keyframe edits across time.

%% file: sec/6_sup.tex
\clearpage

\appendix

\setcounter{figure}{0}
\renewcommand{\thefigure}{S\arabic{figure}}
\setcounter{table}{0}
\renewcommand{\thetable}{S\arabic{table}}
\setcounter{equation}{0}
\renewcommand{\theequation}{S\arabic{equation}}

\section{Appendix Overview}
In this appendix, we provide additional details and results that are not included in the main paper
due to the space limit. The attached video includes intuitive and interesting qualitative results of
V-RGBX.

\section{Workflow \& Implementation Details}
\subsection{Video editing workflow explanation}
\begin{figure*}[t]
  \centering
  \includegraphics[width=\linewidth]{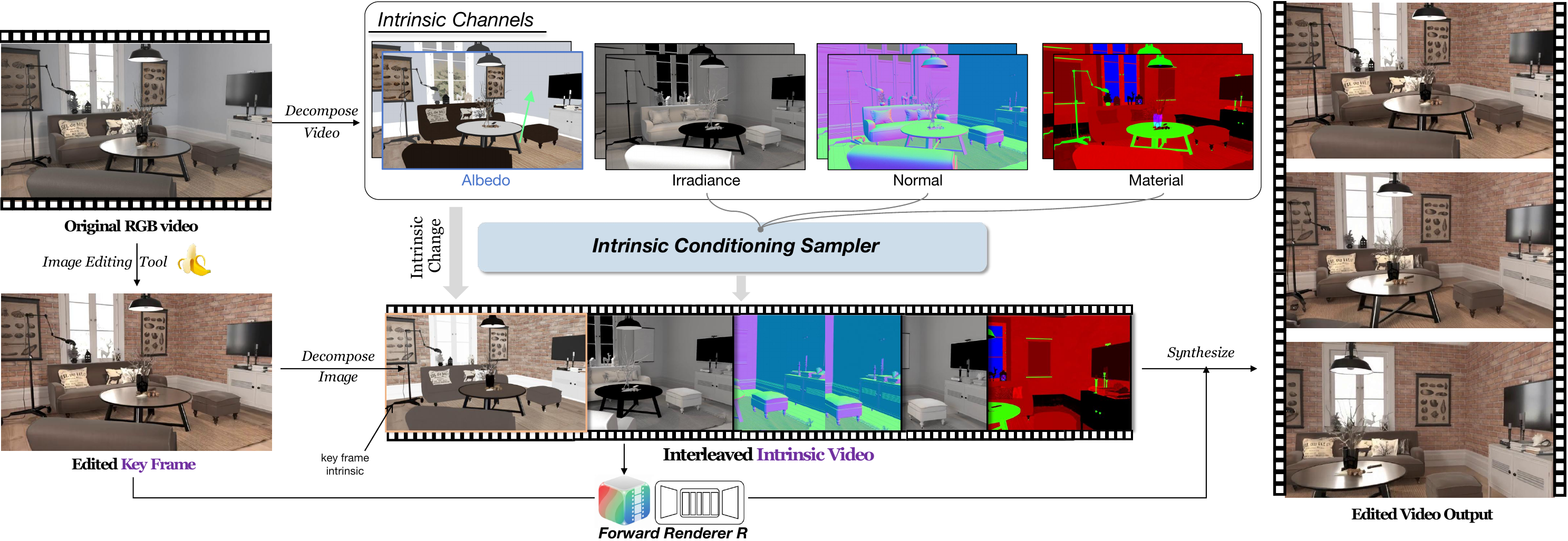}
  \caption{\textbf{Intrinsic-aware video editing workflow of V-RGBX.} Given an input video and edited keyframes, we decompose them into intrinsic channels, and the intrinsic conditioning sampler uses these representations to produce an intrinsic video. The forward renderer then synthesizes the final edited sequence using both the intrinsic video and the appearance cues provided by the edited keyframes.}
  \label{fig:edit_workflow}
\end{figure*}

\paragraph{Intrinsic decomposition and keyframe editing.}
As shown in Fig.~\ref{fig:edit_workflow}, we first decompose the input RGB video into intrinsic channels, including albedo, irradiance, normal, and material.
These intrinsic channels form a physically structured representation that separates appearance, illumination, and geometry, enabling more reliable and controllable video editing.
Selected keyframes are edited with a text-driven image editing tool NanoBanana and then decomposed again to obtain their edited intrinsics, ensuring that user-intended modifications (e.g., material changes or relighting) are reflected in the intrinsic domain.

\vspace{-3mm}
\paragraph{Intrinsic conditioning sampling.}
To propagate the edits beyond the keyframes, we employ an intrinsic conditioning sampler that aggregates both the original per-frame intrinsics and the edited intrinsic channels.
The sampler constructs an interleaved intrinsic sequence $V_{X}'$, inserting edited intrinsic cues at the keyframe positions while preserving unmodified channels for all other frames.
This provides a unified intrinsic sequence that encodes both preserved and edited content in a temporally aligned manner.

\vspace{-3mm}
\paragraph{Forward rendering of edited content.}
The interleaved intrinsic video is then passed through our forward renderer $R$, which synthesizes the final edited RGB video.
The edited keyframes provide both edited intrinsic cues and reference appearance keyframes, and conditioning on intrinsics leverages their structured, disentangled nature to support faithful and controllable propagation of edits. We show more qualitative results in Sec.~\ref{sup:more_edit} and \ref{sup:real}.


\begin{figure}[t]
  \centering
  \includegraphics[width=\linewidth]{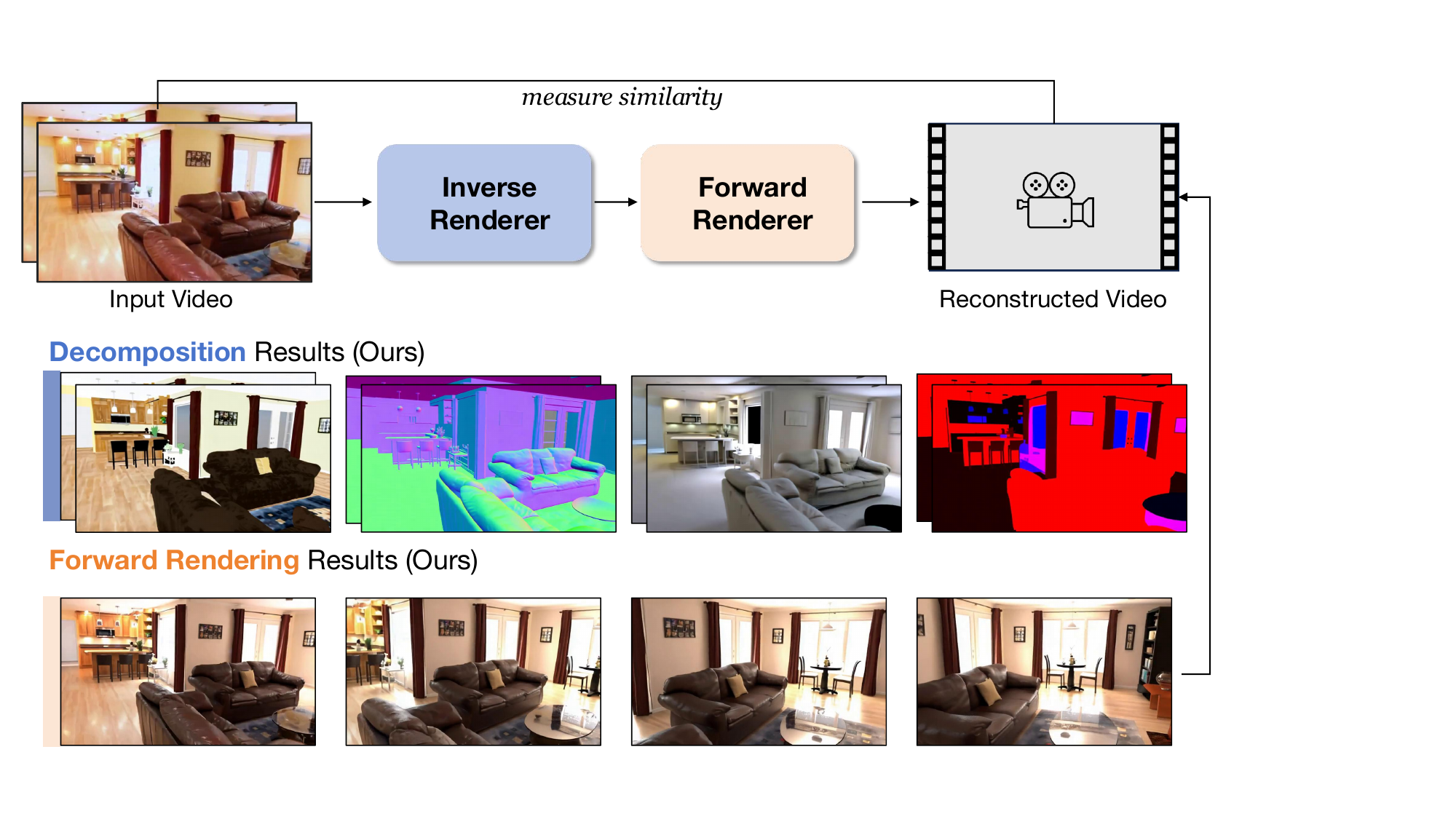}
  \caption{\textbf{Overview of RGB→X→RGB cycle workflow.} An input RGB video is first decomposed into intrinsic components by our inverse renderer, then reconstructed by the forward renderer using the predicted intrinsic sequence and a first-frame keyframe. The decomposition and forward-rendering results illustrate the quality of our intrinsic predictions and the rendered video.}
  \label{fig:cycle_workflow}
\end{figure}

\begin{figure}[t]
  \centering
  \includegraphics[width=\linewidth]{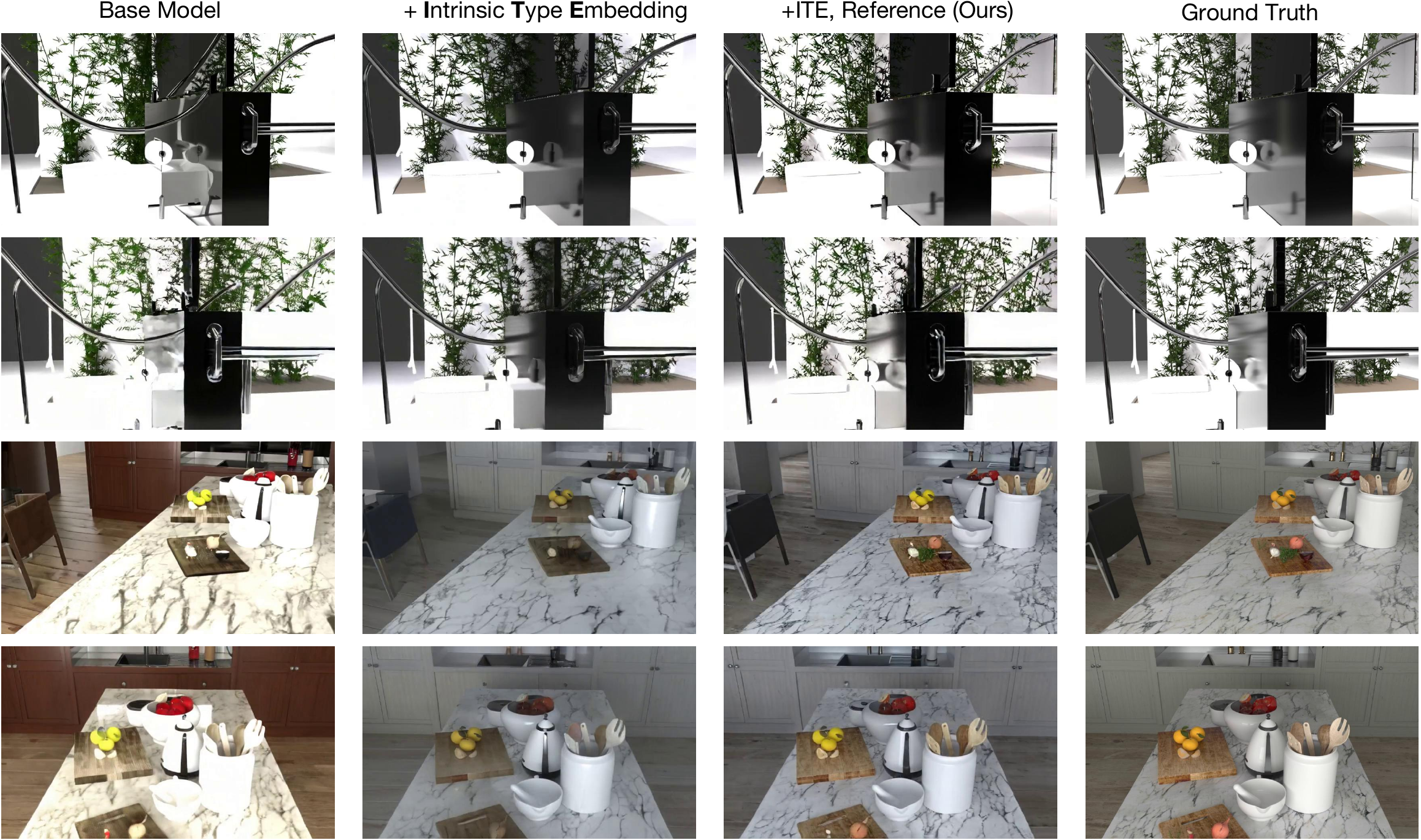}

  \caption{\textbf{Visual ablations on the Intrinsic Type Embedding (ITE) module and the reference condition.} Columns show the base model, adding ITE, adding both ITE and the reference condition (ours), and the ground truth. The intrinsic maps (top) and reconstructed RGB frames (bottom) illustrate that ITE reduces temporal and modality inconsistencies, while the reference condition further improves reflections and color fidelity.}
  \label{fig:ablation}
\end{figure}

\subsection{RGB→X→RGB cycle workflow}
We adopt an RGB→X→RGB cycle setup to assess how well the intrinsic representation retains the information needed for accurate reconstruction and for supporting reliable edit propagation. This evaluation setting provides a clear and comprehensive way to examine how appearance, geometry-related cues, and illumination are preserved when passing through each stage of our framework.

As illustrated in Fig.~\ref{fig:cycle_workflow}, an input RGB video is first decomposed by our inverse renderer into its intrinsic channels. The predicted intrinsic sequence is temporally consistent, and the intrinsic output of the first frame additionally serves as a keyframe to anchor the forward synthesis. These intrinsic channels are then fed into our forward renderer to reconstruct the video. By comparing the reconstructed sequence with the original input, as reported in Table 3, we evaluate how well the intrinsic space maintains pixel-level fidelity, structural detail, and temporal continuity with baseline methods. This cycle analysis also indicates the stability of intrinsic-based edits when propagated across frames. We show more qualitative results in Sec~\ref{sup:more_cycle}.

\subsection{Inference details}
During inference of forward rendering, classifier-free guidance is applied to the reference branch while keeping $V_X'$ as a shared condition:
\begin{equation}
\label{eq:cfg}
\epsilon_{\mathrm{CFG}}
=\epsilon_\theta(z_t, \varnothing, V_X')
+s\!\left[\epsilon_\theta(z_t, v_{\text{ref}}, V_X')-\epsilon_\theta(z_t, \varnothing, V_X')\right],
\end{equation}
where the two terms denote predictions without/with the reference input, respectively. Following the notation in the main text, the reference is defined as $v_{\mathrm{ref}}=\{v'_{i_1},\ldots,v'_{i_k}\}$. This modulates reference-driven appearance while preserving the structural/physical priors encoded in $V_X'$. In our implementation, the guidance scale is set as $s = 1.5$.


\section{Additional Experiments}
\label{sec:exp}

In this section, we present additional ablation studies focusing on the two modules that most strongly affect the propagation behavior of our Forward Renderer: the Intrinsic Type Embedding (ITE) module and the Reference Condition.
For each ablation, we remove the corresponding module and retrain the model from scratch under the same training iterations and hyperparameters as V-RGBX. Both quantitative and qualitative analyses are provided.




\subsection{The Effectiveness of ITE Module}
As shown in Tab.~\ref{tab:train_ablation}, comparing the first and second rows reveals that removing the ITE module—and thus relying solely on interleaved intrinsic conditioning—leads to consistent drops across all evaluation metrics. The qualitative results in Fig.~\ref{fig:ablation} further show noticeable temporal flickering and color inconsistencies.
We observe that when intrinsic channels are interleaved on a per-frame basis without explicit type disambiguation, the model may confuse modality identities across frames. Such confusion can couple signals across different channels in the color space, causing visually incorrect or unstable predictions.

After introducing the ITE module, all metrics improve, and the generated videos (Fig.~\ref{fig:ablation}, second column) exhibit much better temporal stability and appearance consistency. This confirms that ITE provides an effective mechanism for disentangling the intrinsic channels over time, reducing cross-channel conflicts and producing more reliable visual outcomes.

\subsection{The Effectiveness of Reference Condition}
As shown in Tab.~\ref{tab:train_ablation}, adding the reference condition leads to further improvements over using ITE alone.
Note that this experiment evaluates whether the model is trained with reference supervision, which differs from the evaluation in the main paper where we study whether reference images are provided at inference time. Here, the ablation aims to understand the contribution of the reference module itself.

The qualitative comparisons also show consistent gains: in the first two rows of Fig.~\ref{fig:ablation}, reflections become clearer and more coherent, while in the third and fourth rows, object colors and tones align more closely with the ground truth.
These observations suggest that the reference condition offers important complementary cues that help correct biases in the X→RGB mapping and significantly improve reconstruction fidelity.

\begin{table*}[t]
\centering
\caption{\textbf{Quantitative ablations of the ITE module and the reference condition.} 
Adding ITE consistently improves reconstruction quality and temporal stability across all metrics, and further incorporating the reference condition yields the best overall performance, with noticeable gains in PSNR, LPIPS, FID, FVD, and smoothness.}
\vspace{3pt}
\begin{tabular}{lcccccccc}
\toprule
Method & ITE & Key Reference & PSNR$\uparrow$ & LPIPS$\downarrow$ & SSIM$\uparrow$ & FID$\downarrow$ & FVD$\downarrow$ & Smooth.$\uparrow$ \\
\midrule
Base (no added modules)      
& $\times$ & $\times$ 
& 20.96 & 0.2372 & 0.7818 & 37.81 & 532.21 & 0.9769 \\
+ ITE                  
& $\checkmark$ & $\times$
& 21.47 & 0.2149 & \textbf{0.7994} & 35.39 & 405.79 & 0.9802 \\
+ ITE + Reference (ours) 
& $\checkmark$ & $\checkmark$
& \textbf{22.42} & \textbf{0.1930} & 0.7952 & \textbf{29.83} & \textbf{367.89} & \textbf{0.9805} \\
\bottomrule
\end{tabular}
\label{tab:train_ablation}
\end{table*}





\section{Additional Qualitative Results}
\label{sec:results}

\subsection{Additional RGB→X Results}
\begin{figure*}[t]
  \centering
  \includegraphics[width=0.95\linewidth]{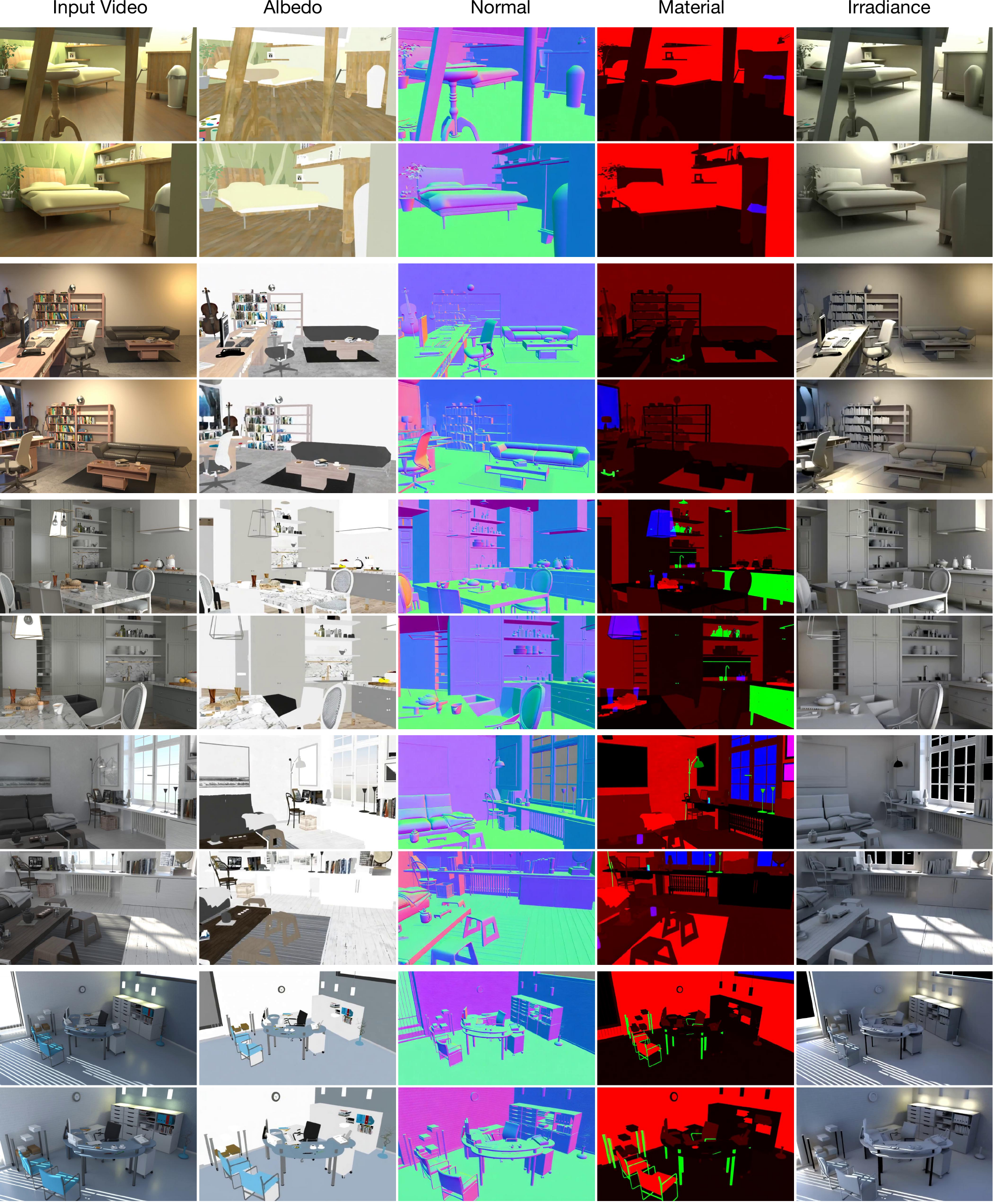}
  \caption{\textbf{RGB→X results on synthetic Evermotion scenes.} Given an input RGB video, V-RGBX decomposes it into albedo, normal, material, and irradiance channels. Each pair of rows shows two frames from the same video, and the second to fifth columns visualize the corresponding intrinsic channels, demonstrating spatially coherent and temporally stable decompositions across diverse indoor scenes.}
  \label{fig:rgb2x_ever}
  \vfill
\end{figure*}

\begin{figure*}[t]
  \centering
  \includegraphics[width=0.95\linewidth]{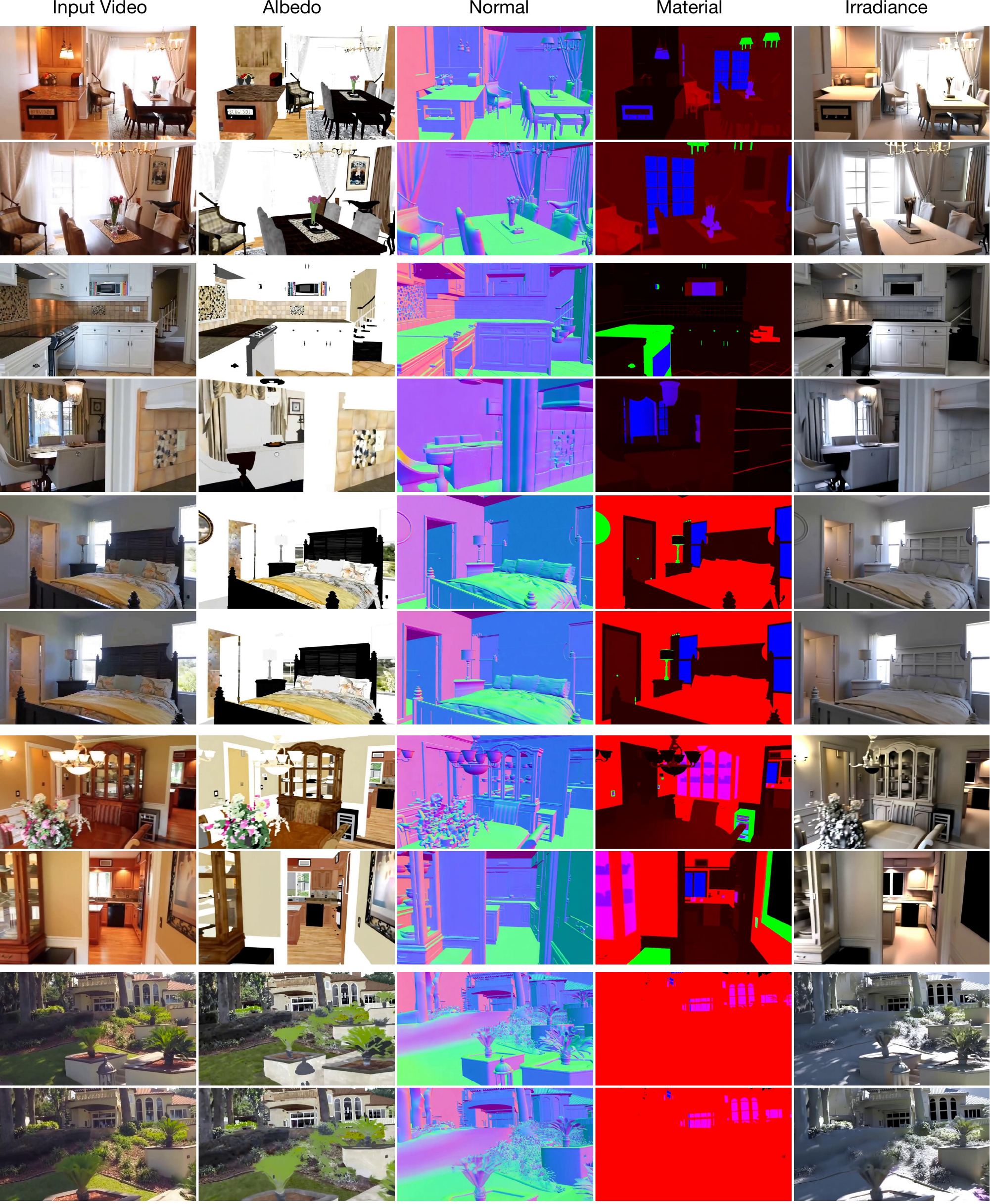}
\caption{\textbf{RGB→X results on real-world RealEstate10K videos.} Given an input RGB video, V-RGBX decomposes it into albedo, normal, material, and irradiance channels. Each pair of rows shows two frames from the same video, and the second to fifth columns visualize the corresponding intrinsic channels, demonstrating coherent and temporally stable decompositions under challenging and unseen real-world conditions, while also showing reasonable generalization to outdoor scenes.}
\label{fig:rgb2x_real}
  \vfill
\end{figure*}

In the main paper (Sec.~4.2), we have already reported quantitative results for the inverse-rendering task (RGB→X). Here, we provide additional qualitative results in Fig.~\ref{fig:rgb2x_ever} and Fig.~\ref{fig:rgb2x_real}, along with representative comparisons against baseline methods.

\begin{figure*}[t]
  \centering
  \includegraphics[width=0.98\linewidth]{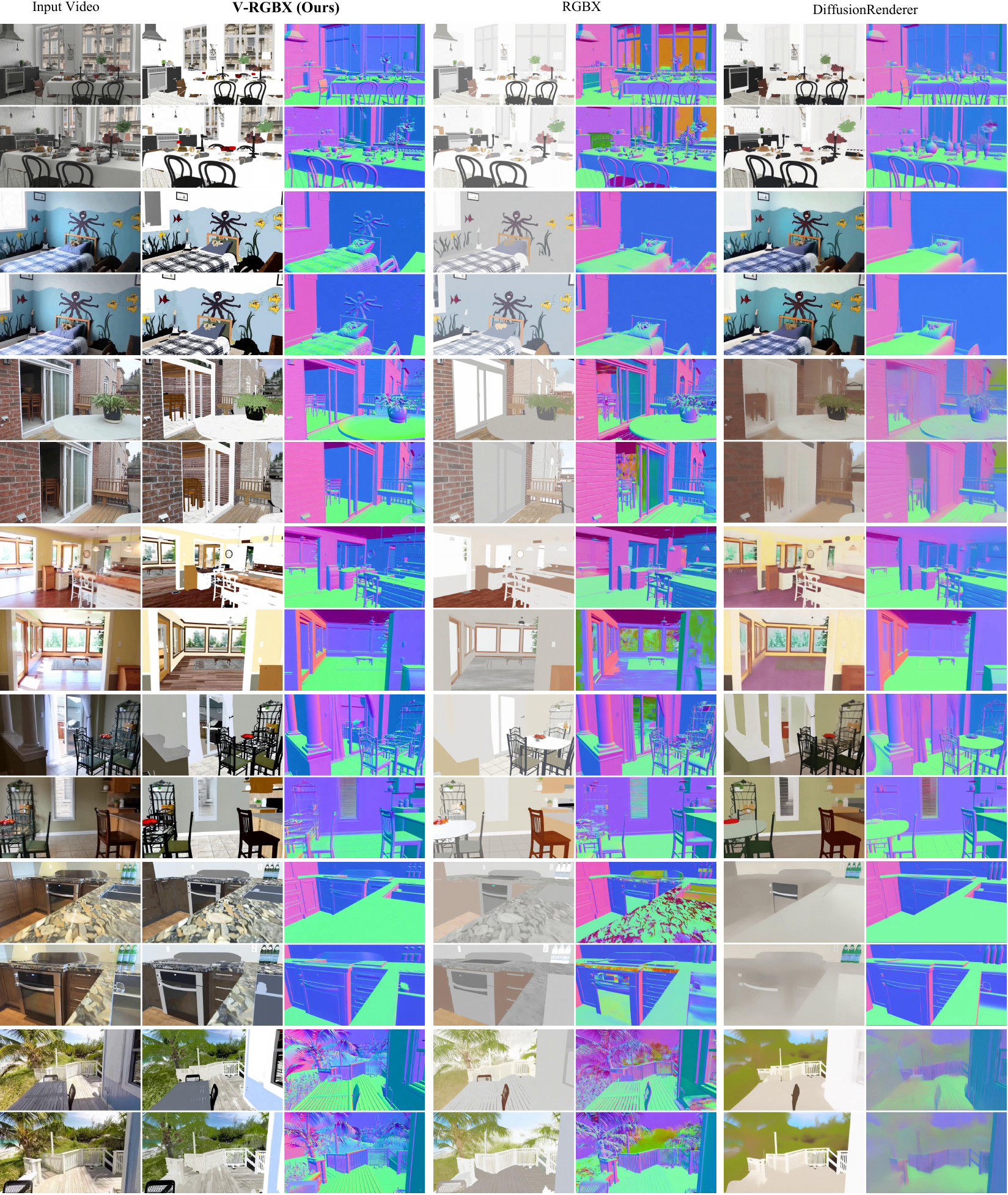}
  \caption{\textbf{Comparison of RGB→X decomposition results with baselines.} Each pair of rows shows two frames from the same input video (first column). For each method, the two columns visualize the predicted albedo and normal channels. Compared with RGBX and DiffusionRenderer, V-RGBX produces intrinsic decompositions with higher visual fidelity, more accurate albedo estimation, and more consistent normal predictions across frames.}
  \label{fig:rgb2x_comp_an}
  \vfill
\end{figure*}

Fig.~\ref{fig:rgb2x_ever} and Fig.~\ref{fig:rgb2x_real} show input videos from both synthetic and real scenes together with the intrinsic predictions produced by V-RGBX, including albedo, normal, material, and irradiance channels. We further compare our method with RGBX and DiffusionRenderer in Fig.~\ref{fig:rgb2x_comp_an}. Consistent with the quantitative findings, V-RGBX yields more stable albedo and normal reconstructions, while RGBX often exhibits temporal instability and color inconsistencies. DiffusionRenderer also shows some failure cases, such as collapsed normal maps and inaccurate color estimates. Moreover, our model demonstrates strong generalization ability, producing reliable intrinsic decompositions even under challenging real-world and outdoor lighting conditions.


\subsection{Additional X→RGB Results}

\begin{figure*}[t]
  \centering
  \vspace{-6pt}
  \includegraphics[width=0.92\linewidth]{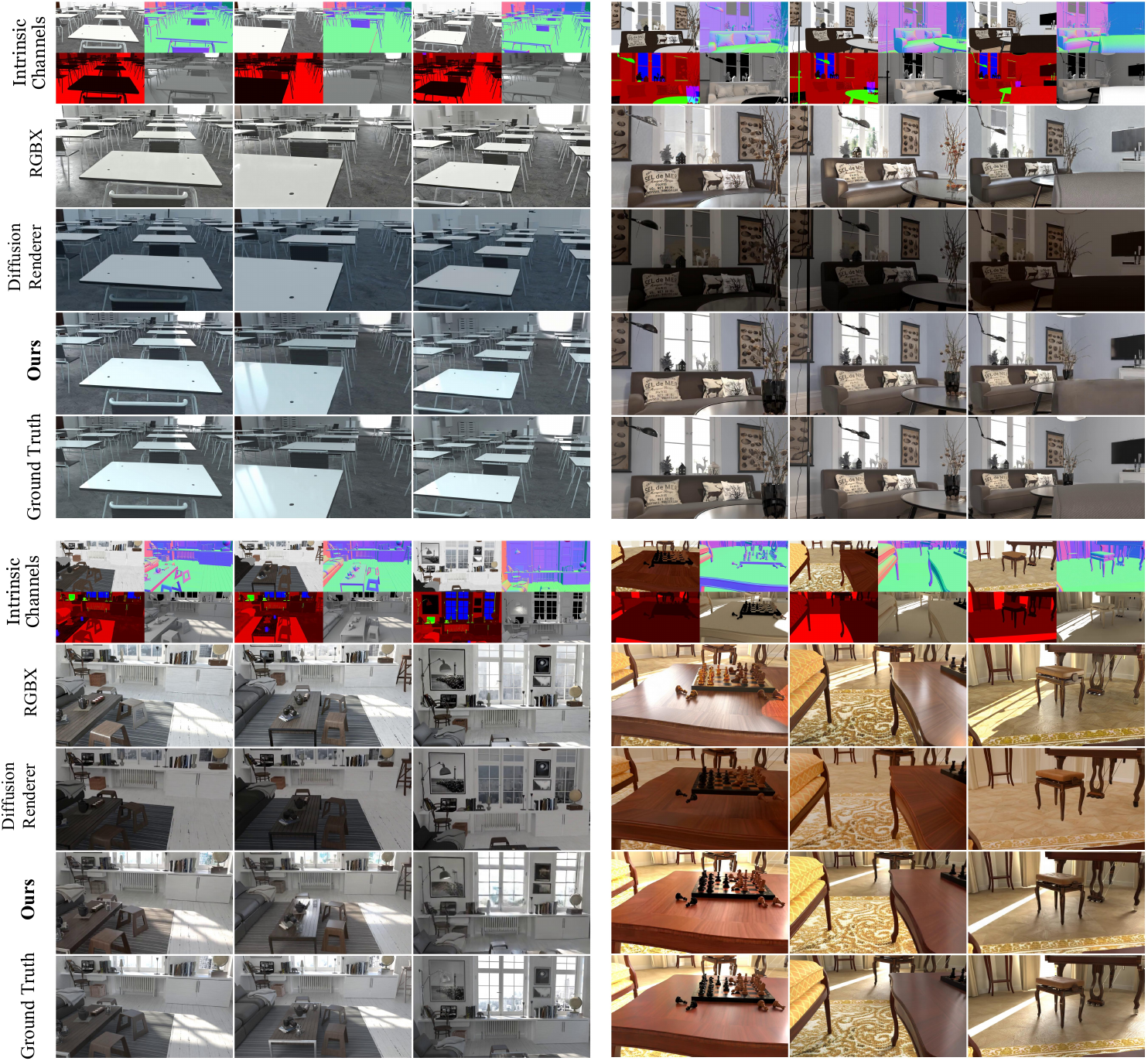}
\caption{\textbf{More qualitative comparisons for the X→RGB task.} Each group of three columns shows three frames from the same video, while each row corresponds to a different method: intrinsic channels inputs, RGBX, DiffusionRenderer, our results, and the ground truth. The comparisons illustrate that our method performs better in scene appearance and temporal consistency across frames.}

  \label{fig:more_x2video}
  \vfill
\end{figure*}

As discussed in Sec.~4.3, we evaluate the X→RGB task, and Fig.~\ref{fig:more_x2video} provides additional qualitative examples together with comparisons against the baseline methods. These examples show that V-RGBX handles complex lighting effects and geometric structures more reliably, producing RGB sequences with more stable shading, reflections, and temporal coherence. Overall, the supplemental results further illustrate the robustness of our approach when generating videos from intrinsic representations.

\subsection{Additional RGB→X→RGB Results}
\label{sup:more_cycle}

\begin{figure*}[t]
  \centering
  \includegraphics[width=0.96\linewidth]{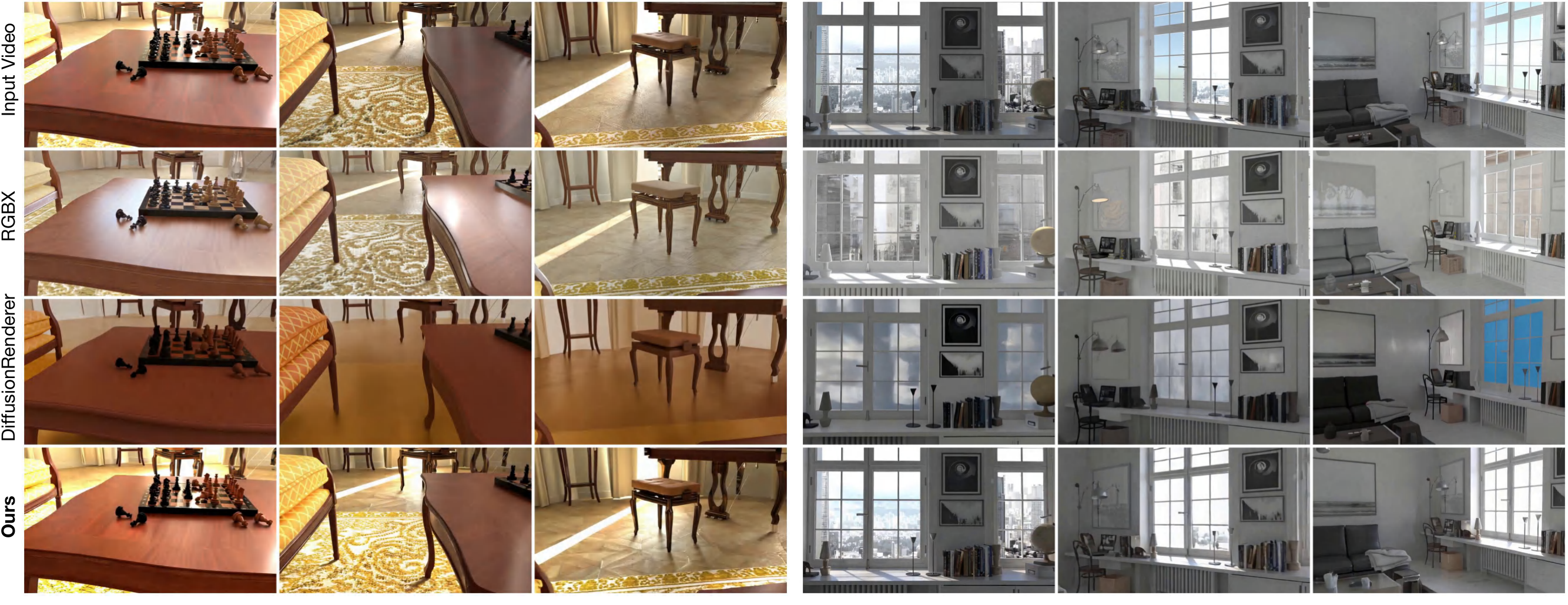}
  \caption{\textbf{RGB→X→RGB cycle results on the synthetic dataset.} Each row shows a different method (the first row is the input video as ground truth). Every three columns correspond to three frames from the same video. Our method produces reconstructions closest to the ground truth and better preserves scene appearance and structure throughout the sequence.}
  \label{fig:cycle_ever}
  \vfill
\end{figure*}

\begin{figure*}[t]
  \centering
  \includegraphics[width=0.96\linewidth]{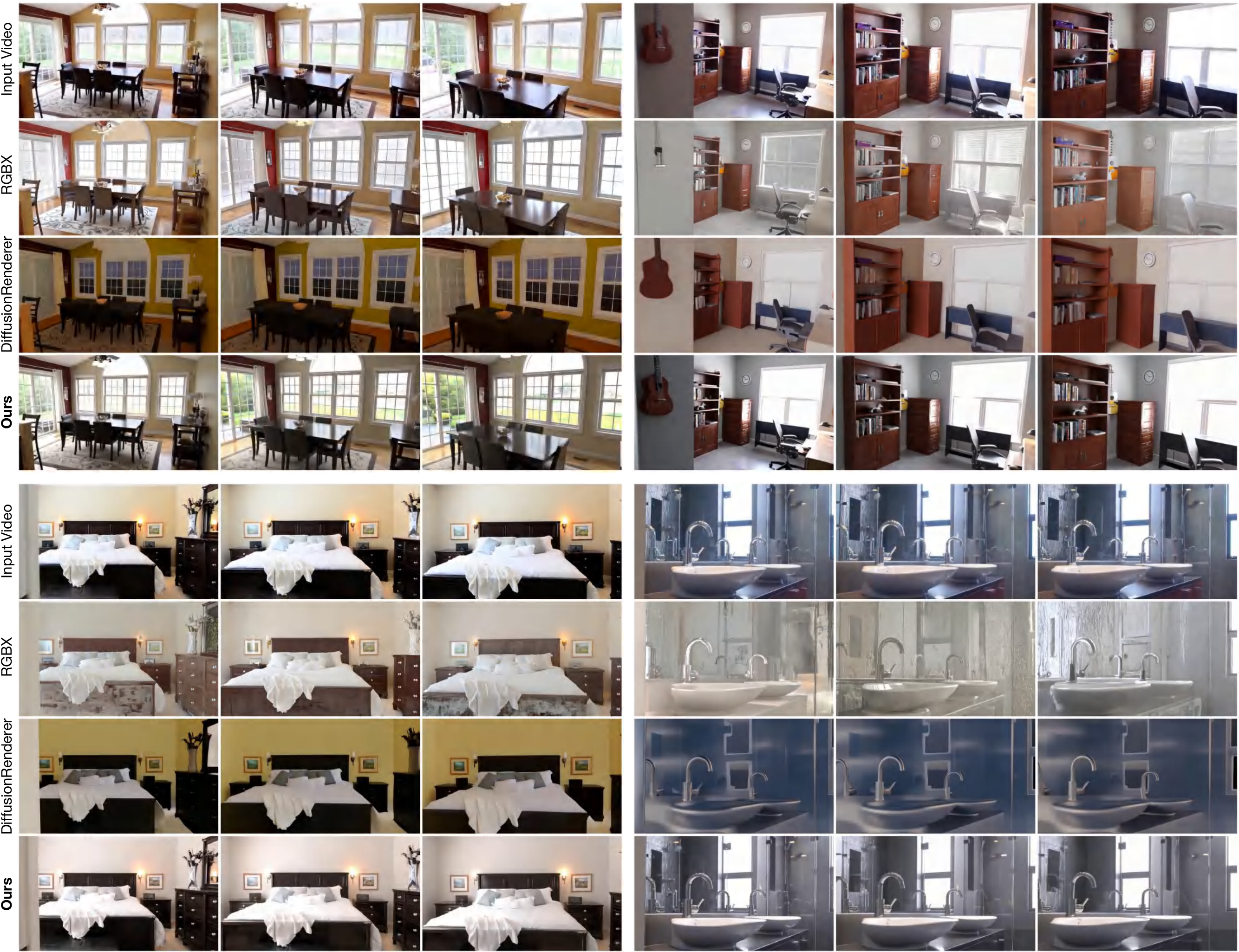}
  \caption{\textbf{RGB→X→RGB cycle results on the real-world dataset.} Each row shows a different method (the first row is the input video as ground truth). Every three columns correspond to frames from the same video. Our method gives a closer match to the ground truth.}
  \label{fig:cycle_real}
  \vfill
\end{figure*}

As discussed in Sec.~4.3, we quantitatively evaluate the RGB→X→RGB cycle to assess whether the intrinsic representation preserves sufficient information for accurate reconstruction and reliable edit propagation. In Figs.~\ref{fig:cycle_ever} and~\ref{fig:cycle_real}, we provide additional qualitative comparisons on both synthetic and real-world videos, showing the reconstructed sequences produced by our approach and the baseline methods. Our method achieves more stable temporal behavior and better preserves scene appearance across the full cycle.

\subsection{Keyframe Editing Results}
\label{sup:more_edit}

\begin{figure*}[t]
  \centering
  \includegraphics[width=\linewidth]{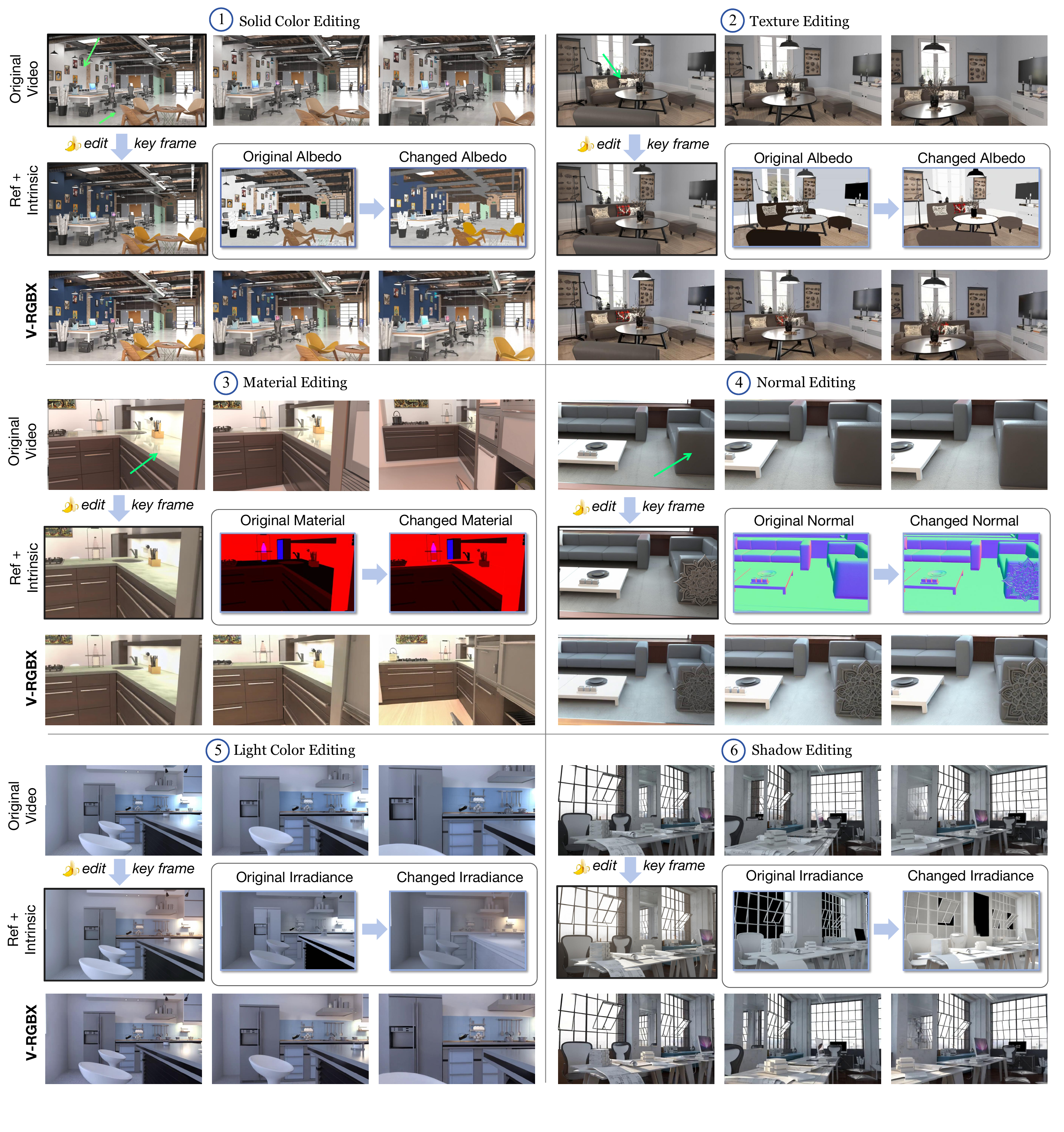}
   \caption{\textbf{Intermediate results of the intrinsic-aware keyframe editing.} We visualize intermediate results used by V-RGBX across the following editing types: (1) solid color, (2) texture, (3) material, (4) normal, (5) light color, and (6) shadow editing. For each case, we show the original video frames, the edited keyframe produced by the NanoBanana tool, and the corresponding modified intrinsic channels (albedo, material, normal, or irradiance) that serve as conditioning inputs. These processes reveal how keyframe edits are translated into intrinsic-space modifications, which are then reliably propagated by V-RGBX to generate the final temporally consistent edited video.}
  \label{fig:sup_edit}
\end{figure*}

As described in Sec.~4.5, we demonstrate the intrinsic-aware video editing capability of V\textminus RGBX in the main paper.
To provide a clearer view of how the edits are propagated through our intrinsic pipeline, we include the complete set of intermediate results in Fig.~\ref{fig:sup_edit}. Specifically, we visualize the input video frames, the edited keyframes produced by the NanoBanana tool, and the extracted intrinsic channels that jointly condition the generation process. These intermediate visualizations help illustrate how the edited albedo, normal, material, or irradiance attributes guide the final synthesis. Following the editing workflow shown in Fig.~\ref{fig:edit_workflow}, V-RGBX takes the modified keyframes and intrinsic channels as conditioning signals and generates temporally consistent, intrinsically coherent outputs. 





\subsection{Real-world Challenging Cases}
\label{sup:real}

We provide additional demonstrations of our editing capability on diverse and challenging real-world scenarios. Please refer to the attached video for full results. Our evaluations cover real indoor scenes, self-captured videos, general object videos, and cases with complex lighting, showcasing robust intrinsic-aware editing performance across a wide range of real-world conditions.



\clearpage

\begin{figure*}[t]
  \centering
  \includegraphics[width=0.95\linewidth]{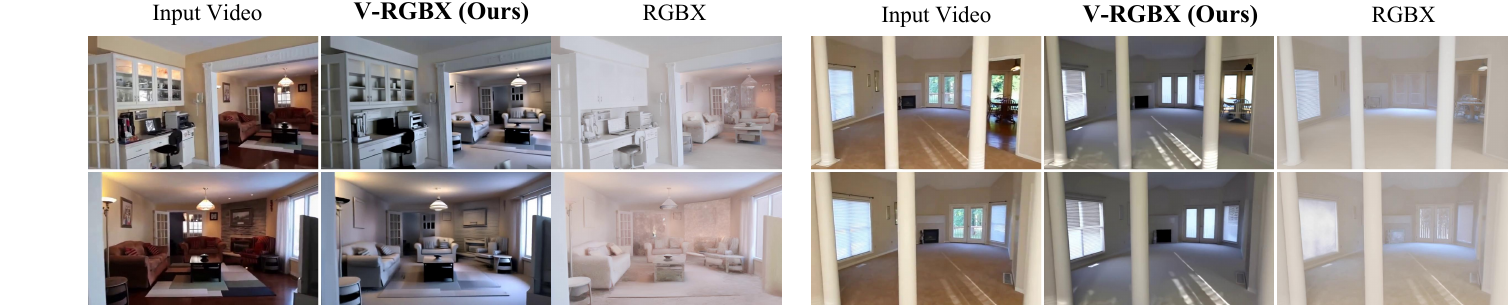}
  \caption{\textbf{Comparison of irradiance decomposition with baselines.} The figure shows two different videos, with each pair of rows representing two frames from the same video. For each frame, the second and third columns show irradiance predictions from V-RGBX and RGBX. V-RGBX produces more accurate illumination and shadow modeling, resulting in clearer and more plausible irradiance maps.}
  \label{fig:rgb2x_comp_i}
\end{figure*}

